\documentclass[10pt,twocolumn,letterpaper]{article}

\usepackage[pagenumbers]{cvpr} 
\definecolor{cvprblue}{rgb}{0.21,0.49,0.74}
\usepackage[pagebackref,breaklinks,colorlinks,allcolors=cvprblue]{hyperref}
\usepackage{booktabs}
\usepackage{multirow}
\usepackage{subcaption}
\usepackage{colortbl}

\title{MV-SSM: Multi-View State Space Modeling for 3D Human Pose Estimation}

\author{%
Aviral Chharia$^{1}$, Wenbo Gou$^{1}$, Haoye Dong$^{2, }$\thanks{Corresponding author}\\
$^1$Carnegie Mellon University \ $^2$National University of Singapore\\
{\tt \{achharia, wgou\}@andrew.cmu.edu, haoye@nus.edu.sg}\\
}
\begin{document}
\maketitle
\begin{abstract}
While significant progress has been made in single-view 3D human pose estimation, multi-view 3D human pose estimation remains challenging, particularly in terms of generalizing to new camera configurations. Existing attention-based transformers often struggle to accurately model the spatial arrangement of keypoints, especially in occluded scenarios. Additionally, they tend to overfit specific camera arrangements and visual scenes from training data, resulting in substantial performance drops in new settings. In this study, we introduce a novel \textbf{M}ulti-\textbf{V}iew \textbf{S}tate \textbf{S}pace \textbf{M}odeling framework, named \textbf{MV-SSM}, for robustly estimating 3D human keypoints. We explicitly model the joint spatial sequence at two distinct levels: the feature level from multi-view images and the person keypoint level. We propose a Projective State Space (PSS) block to learn a generalized representation of joint spatial arrangements using state space modeling. Moreover, we modify Mamba's traditional scanning into an effective Grid Token-guided Bidirectional Scanning (GTBS), which is integral to the PSS block. Multiple experiments demonstrate that MV-SSM achieves strong generalization, outperforming state-of-the-art methods: {\color{red} $\mathbf{+10.8}$} on AP$_{25}$ {\color{blue} $(+24\%$$\uparrow)$} on the challenging three-camera setting in CMU Panoptic, {\color{red} $\mathbf{+7.0}$} on AP$_{25}$ {\color{blue} $(+13\%$$\uparrow)$} on varying camera arrangements, and {\color{red} $\mathbf{+15.3}$} PCP {\color{blue} $(+38\%$$\uparrow)$} on Campus A1 in cross-dataset evaluations. Project Website:~\url{https://aviralchharia.github.io/MV-SSM}.
\end{abstract}
\vspace{-1em}
\section{Introduction}
\label{sec:intro}

Estimating 3D human pose from multiple views is a crucial task in computer vision, with diverse applications in sports, industrial monitoring, and surveillance.
However, it remains challenging due to its inherent complexity. Unlike single-view methods~\cite{akhter2015pose,fan2014pose,ramakrishna2012reconstructing,wang2014robust,yasin2016dual,martinez2017simple,dabral2018learning,Li_2022_CVPR}, estimating 3D human pose from multi-view inputs involves several challenges, including---keypoint-level association to find cross-view correspondences while fusing multi-view information, multi-person association, and high occlusion.

\begin{figure}
    \centering
    \includegraphics[width=0.49\textwidth]{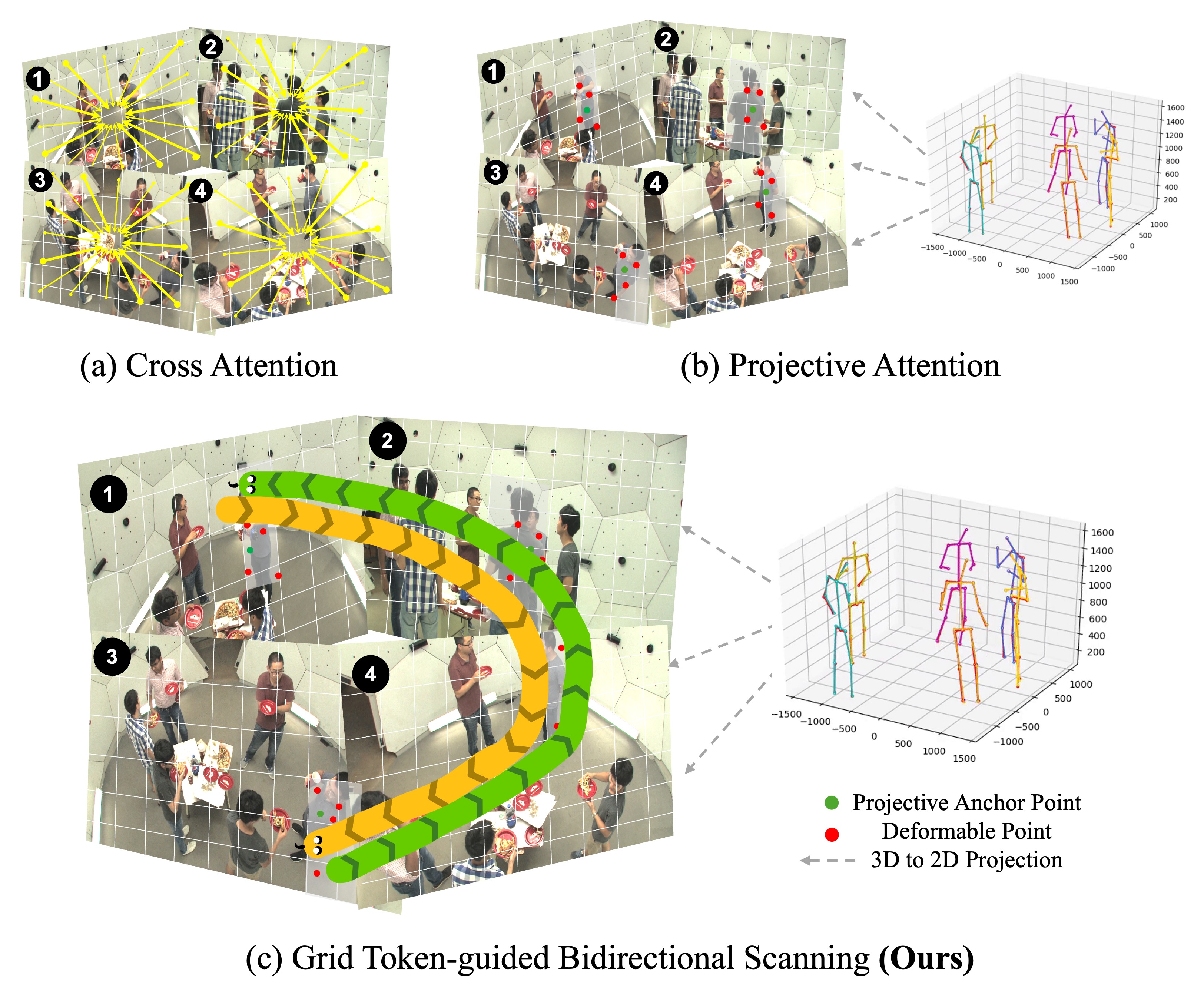}
    \caption{\textbf{Motivation.} Comparison of different token scanning methods. (a) Cross Attention acts on all image tokens. (b) Projective Attention~\cite{zhang2021direct_mvp, liao2024multiple_mvgformer} obtains anchors with perspective projection and selectively attends to sample tokens surrounding the anchor points. (c) The proposed Grid Token-guided Bidirectional Scanning (GTBS) encodes the local context and the joint spatial sequence at the visual feature and person-keypoint levels.}
    \label{fig:motivation}
\end{figure}

Previous works~\cite{dong2019fast,chu2021part,chen2020cross,bridgeman2019multi,chen2020multi,dong2021fast,perez2022matching,zhang20204d} have tried to address this task using a multi-stage approach. They first estimated the 2D keypoints in all views using off-the-self models such as OpenPose~\cite{cao2019openposerealtimemultiperson2d} and YOLOv10~\cite{wang2024yolov10}. Next, they associate these keypoints with their respective human subjects before finally triangulating the matched 2D keypoints to obtain 3D estimates.
Consequently, recent works have focused on refining individual steps in these multi-stage frameworks. However, the multi-stage paradigm has limitations, as the matching algorithm is susceptible to complications in scenes with multiple humans, and errors tend to accumulate at each stage of the pipeline. To address this, a few recent works~\cite{iskakov2019learnable,reddy2021tessetrack,tu2020voxelpose,ye2022faster,zhang2021direct_mvp,liao2024multiple_mvgformer} have focused on end-to-end supervised learning approaches. These models bypass the explicit geometry modeling required in multi-stage models. However, they suffer from a lack of generalizability to new scenes and camera arrangements due to overfitting to training camera setups.

To tackle this challenge, we introduce \textit{MV-SSM}, a novel Multi-View State Space Modeling framework. \textit{MV-SSM} is the \textit{first} framework that integrates state space models~\cite{gu2023mamba} into 3D multi-view geometry modeling for 3D human pose estimation. By explicitly modeling the joint spatial sequences at both the feature level from multi-view images and the person keypoint level, \textit{MV-SSM} harnesses the sequence-modeling strengths of state space models to accurately predict 3D human keypoints, even in scenarios with significant occlusions.

\textit{MV-SSM} is designed as an end-to-end differentiable model that progressively refines 3D keypoint predictions using state space modeling blocks. It first extracts multi-scale features from multi-view inputs using a ResNet-50~\cite{he2016deep} backbone. These features are then processed through a carefully designed Projective State Space (PSS) block, which models the spatial relationships between keypoints across different views. The PSS block utilizes projective attention to effectively fuse multi-view information and incorporates SSMs to learn the joint spatial sequence. Through this, \textit{MV-SSM} can handle occlusions and improve the accuracy of 3D keypoint estimation. \textit{MV-SSM} further enhances its predictions by progressively refining the keypoint estimates through successive layers of PSS blocks, which update the tokens and adjust the keypoint positions with residual offsets. Finally, \textit{MV-SSM} predicts the 3D human poses via differentiable algebraic triangulation, ensuring robust and accurate estimation. We summarize our contributions as:

\begin{itemize}
    \item The present study is the \textit{first} to adapt visual mamba for the 3D multi-view human pose estimation task. It presents \textit{MV-SSM}, a novel \textbf{M}ulti-\textbf{V}iew Geometry \textbf{S}tate \textbf{S}pace~\textbf{M}odel, designed to improve the robustness and generalizability of localized 3D skeleton keypoints.
    
    \item Our Projective State Space (PSS) block integrates state space modeling and projection attention to effectively capture joint spatial sequences and fuse multi-view information.
    
    \item We introduce Grid-Token guided Bidirectional Scanning (GTBS) to enhance performance further and validate its effectiveness through ablations.
    
    \item Experiments demonstrate that our \textit{MV-SSM} significantly outperforms SOTA methods on in-domain 3D keypoints estimation and multiple generalizability tests, validating its adaptability to new camera settings and scenarios.
\end{itemize}
\section{Related works}
\label{sec:related_works}

\subsection{3D Human Pose Estimation}
\noindent \textbf{Single-view Approaches.} Estimating 3D human pose from a single image is inherently ill-posed due to challenges of occlusion and depth ambiguity. Current approaches primarily apply two paradigms: \textit{(i)} 2D-to-3D lifting architectures, and \textit{(ii)} direct regression frameworks. Lifting-based schemes~\cite{akhter2015pose,fan2014pose,ramakrishna2012reconstructing,wang2014robust,yasin2016dual,martinez2017simple,dabral2018learning,Li_2022_CVPR} first detect robust 2D keypoints, and then lifts them into corresponding 3D space by depth estimation through geometric constraints and learned priors. Recent advances exploited 3D CNNs and transformers~\cite{vaswani2017attention}, to enable end-to-end 3D regression directly from the image. Li~\textit{et~al}.~\cite{li20153d} introduced dual-task learning combining keypoint detection and regression. Volumetric approaches like~\cite{pavlakos2017coarse} employed stacked ConvNets to estimate the likelihoods of keypoints. Few recent methods~\cite{ma2021context} integrated GNNs~\cite{defferrard2016convolutional} with attention mechanisms to constraint limb length priors and contextual relationships.\\
\vspace{-0.5em}

\noindent \textbf{Multi-view Approaches.} Multi-view methods offer relatively accurate pose estimation, as occluded keypoints in one view are often visible in others. However, developing generalizable learning-based models that effectively utilize multi-view inputs remains a significant challenge. Given multiple camera positions around a scene, 3D keypoints can be estimated via triangulation once 2D keypoints are detected in each view. However, a key challenge in this approach is matching subjects across views. Traditional triangulation-based methods, such as one by Dong~\textit{et al.}~\cite{dong2019fast} achieve high precision with robust pose estimates but suffer from cross-view matching fragility and non-differentiable pipelines. Learning-based alternatives address these issues by integrating differential geometry~\cite{iskakov2019learnable}. One of the first works by Iskakov~\textit{et~al.}~\cite{iskakov2019learnable} introduced volumetric feature unprojection using 3D CNNs for keypoint position estimation. Another widely adopted method, VoxelPose~\cite{tu2020voxelpose}, directly aggregated multi-view features in 3D voxel space. Recent approaches, such as MvP~\cite{zhang2021direct_mvp} and MVGFormer~\cite{liao2024multiple_mvgformer}, leveraged hierarchical querying mechanisms with attention-based fusion, demonstrating improved robustness to occlusions. MVGFormer~\cite{liao2024multiple_mvgformer} introduced a queue to filter out improbable subject candidates and iteratively refine keypoint locations. Due to the additional visual information from multi-view inputs, these methods achieve higher accuracy compared to single-view alternatives.

\subsection{State Space Models} 
State Space Models (SSMs) have evolved from the linear dynamic systems used in Kalman filtering~\cite{kalman1960new} to advanced deep learning implementations. Gu \textit{et al.} introduced Structured State Space Sequence (S4) models~\cite{gu2021s4,gu2021combining} which achieved SOTA performance in capturing long-range dependencies. Mamba~\cite{gu2023mamba} further extended the S4 model by replacing the fixed hidden space projection matrices with a selective projection matrix, dynamically adjusted to match the input sequence. This enhancement demonstrated promising capabilities for processing long-range text sequences.  Emerging works have adapted Mamba for visual tasks. Yang~\textit{et~al.}~\cite{vim} and Liu~\textit{et~al.}~\cite{liu2024vmamba} modified Mamba's scanning mechanism for vision tasks by replacing the original forward scan with bidirectional horizontal scanning of image patches to better capture spatial relationships. Further improvements, such as VMamba~\cite{liu2024vmamba}, introduced enhanced vertical scans, enabling a cross-scan mechanism. Mamba-ND~\cite{li2025mamba} expanded Mamba to video understanding by scanning video frames across height, width, and temporal dimensions from both directions. VideoMamba~\cite{li2025videomamba} followed a vision Transformer (ViT)-inspired design, demonstrating the scalability of SSM models for video processing. Additionally, hybrid Mamba-Transformer approaches have emerged~\cite{hatamizadeh2024mambavisionhybridmambatransformervision}, achieving SOTA performance and high image throughput on ImageNet-1K~\cite{deng2009imagenet} classification. Our proposed MV-SSM is the first to leverage SSMs to the multi-view 3D human pose estimation task, integrating their strengths to improve accuracy and robustness.
\section{Proposed Methodology}

This section first introduces the preliminaries on state space models, followed by an in-depth description of the proposed model architecture, which includes the Projective State Space block and the Grid Token-guided Bidirectional Scanning.

\begin{figure*}[t]
    \centering
    \includegraphics[width=\textwidth]{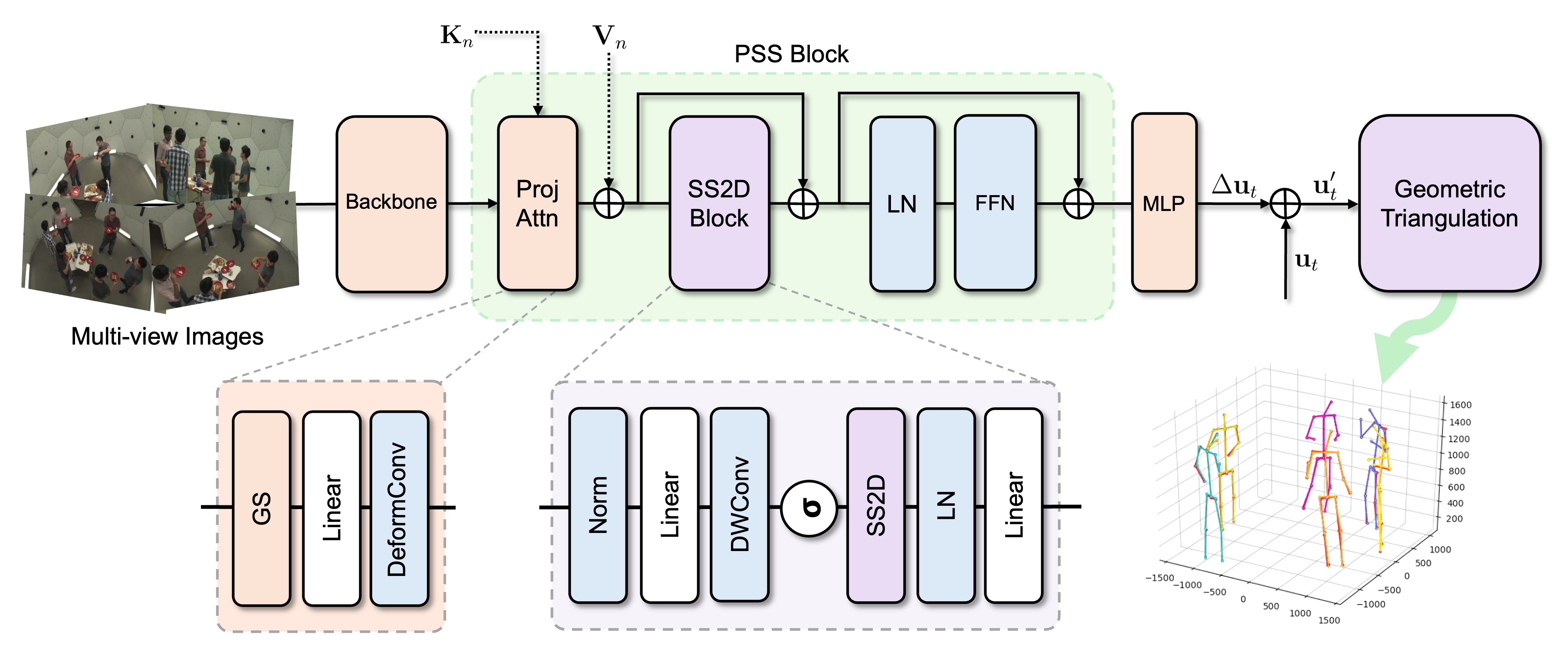}
    \caption{The architecture of the proposed Multi-view State Space Model (\textit{MV-SSM}). It processes multi-view images through a ResNet-50 backbone to extract multi-scale features, which are refined by stacked Projective State Space (PSS) Blocks. These blocks leverage projective attention and state space modeling to progressively refine the keypoints, with final 3D keypoints estimated via geometric triangulation.}
    \label{fig:model_architecture}
    \vspace{-1mm}
\end{figure*}

\subsection{Preliminaries}

\noindent \textbf{SSMs.} Recently, Structured State Space Sequence (S4) models~\cite{gu2021s4} have demonstrated high potential in handling long sequential contexts. State space models transform an input vector $x(t)\in\mathbb{R}$ into an output sequence $ y(t) \in \mathbb{R}$ using latent state $h(t) \in \mathbb{R}^N$. This process follows linear differential equations (as shown in Eq.~\ref{eq:ssm}), where $\boldsymbol{A} \in \mathbb{R}^{N \times N}$, $\boldsymbol{B} \in \mathbb{R}^{N \times 1}$, $\boldsymbol{C} \in \mathbb{R}^{1 \times N}$ and ${D} \in \mathbb{R}^{1}$ form the weight parameters.\\
\begin{minipage}{0.47\textwidth}
\begin{equation}
\label{eq:ssm}
    \begin{split}
  &h'(t) = \boldsymbol{A}h(t) + \boldsymbol{B}x(t),  \\
  &y(t)  = \boldsymbol{C}h(t)+ {D}x(t),
\end{split}
\end{equation}
\end{minipage}
\begin{minipage}{0.47\textwidth}
\vspace{0.5em}
\begin{equation}
\begin{aligned}
    &h_{t} = \overline{\boldsymbol{A}}h_{t-1}+\overline{\boldsymbol{B}}x_{t},\\
    &y_{t} = \boldsymbol{C}h(t).
\end{aligned}\label{discretized}
\end{equation}
\vspace{1mm}
\end{minipage}
\\
The zero-order hold (ZOH) rule is often used to discretize these continuous systems (Eq.~\ref{ZOH}).

\begin{minipage}{0.47\textwidth}
\vspace{1mm}
\begin{equation}
\begin{aligned}
&\overline{\boldsymbol{A}}=\text{exp}(\Delta\boldsymbol{A}), \\&\overline{\boldsymbol{B}}=(\Delta\boldsymbol{A})^{-1}(\text{exp}(\Delta\boldsymbol{A})-\boldsymbol{I})\cdot\Delta\boldsymbol{B} .
\end{aligned}
\label{ZOH}
\vspace{1mm}
\end{equation}
\end{minipage}

\noindent Here, $\Delta$ stands for the discrete step size. The S4 model is a Linear Time Invariant (LTI) system, as the weighting parameters and discretization rules remain fixed over time step $t$. The S6 models~\cite{gu2023mamba} further extend S4 models by adding selective scan, which dynamically adapts the projection matrices along the time dimension of the input sequence. This enables Mamba to project the input into the latent space at each step by remembering and discarding different tokens inside to sequence, compared to the fixed projection matrix of the S4 models.\\
\vspace{-0.5em}

\noindent \textbf{SSMs for Visual tasks.} Mamba~\cite{gu2023mamba} was initially designed for 1D data such as text or audio, which makes its direct application to visual tasks challenging since images contain both local context and global spatial information. To adapt Mamba for visual information, most works first chunk the 2D/ 3D input into small patches and encode them as vectors. Vim~\cite{vim} designed a bidirectional horizontal scan, which goes through the image patches column by column from two directions. VMamba~\cite{liu2024vmamba} proposed an SS2D block with additional vertical scans and introduced it into the VSS block (See the Figure~\ref{fig:pss_block} (a-b)).\\
\vspace{-0.5em}

\noindent \textbf{SSMs for Joint Spatial Sequence.} It is important to note that MV-SSM differs significantly from video-based Mamba, which focuses on learning temporal relationships from consecutive frame sequences. Instead, we model the joint spatial sequence within a series of static multi-view frames. By utilizing Mamba to capture intrinsic keypoint relations in a multi-view context, our method markedly enhances 3D pose estimation accuracy. Specifically in Eq.~\ref{eq:ssm}, $x(t)$ corresponds to the $j^{th}$ joint token of the joint spatial sequence $\mathbf{E}_n$, and our PSS block scans along joint dimension rather than the original time dimension. $y(t)= \text{PSS}(x(t))$ corresponds to the updated token from the PSS block.

\subsection{MV-SSM}

\noindent \textbf{Problem Formulation.} Given a set of multi-view images $\mathbf{I}_t \in \mathbb{R}^{H \times W \times 3}$ for $t = 1,..., T$, captured by $T$ cameras, the goal is to estimate the 3D keypoints for each human present in the captured scene. We formulate this problem as a progressive regression task, where our model~\textit{MV-SSM} progressively estimates and refines its predictions through PSS blocks, ultimately achieving accurate 3D human pose based on multi-view images.\\
\vspace{-0.75em}

\noindent \textbf{Model Architecture.} As shown in Figure~\ref{fig:model_architecture}, \textit{MV-SSM} first passes the multi-view images into a ResNet-50 backbone~\cite{he2016deep} to extract multi-scale features $\{\mathbf{F}_t\}_{t=1}^T$. The extracted features are subsequently processed by our proposed Projective State Space (PSS) block, which leverages the joint spatial sequence by capturing their relationships within state space through GTBS. Finally, for estimating the 3D human keypoints, it employs a learnable triangulation adopted from~\cite{iskakov2019learnable}. We additionally process keypoint-level information through stacked PSS blocks, with each layer outputting joint offsets to iteratively refine the predictions from the previous layer. The final model consists of multiple PSS blocks stacked layer-wise, which enables \textit{MV-SSM} to learn a progressive regression yielding accurate 3D human pose estimation.\\
\vspace{-0.75em}

\noindent \textbf{Joint Token Representation.} Following prior works~\cite{zhang2021direct_mvp,liao2024multiple_mvgformer}, we use a joint token to represent each human in the scene. Each joint token is denoted as $\{\mathbf{E}_n\}_{n=1}^{N}$, where $N$ is the initial set of humans, and represented as a combination of its visual feature term $\mathbf{V}_n \in \mathbb{R}^{J \times L}$, and its geometry term $\mathbf{K}_n \in \mathbb{R}^{J \times 3}$, where $J$ stands for the number of keypoints and $L$ denotes the feature dimension. These two terms form the joint token denoted by $\mathbf{E}_n=\left(\mathbf{V}_n, \mathbf{K}_n \right)$. This representation is beneficial as it explicitly encodes both visual and geometric information. 

While learning the visual feature term, we model two sub-tokens, i.e., person-level $\{\mathbf{s}_n\}_{n=1}^{N} \subset \mathbb{R}^L$ and keypoint-level $\{\mathbf{g}_j\}_{j=1}^{J} \subset \mathbb{R}^L$ sub-tokens. The visual feature for the $j^{th}$ keypoint of the $n^{\text{th}}$ token is represented by $\mathbf{v}_n^j=\mathbf{s}_n+\mathbf{g}_j$. We represent the combined visual feature term of all keypoints as $\mathbf{V}_n=[\mathbf{v}_n^1,\mathbf{v}_n^2,...,\mathbf{v}_n^J]^\intercal$. The geometry term directly stores each keypoint position, i.e., $\mathbf{k}_n^j = [\mathbf{x}_n^j, \mathbf{y}_n^j, \mathbf{z}_n^j]^\intercal$, where keypoint $j$ of the $n^{\text{th}}$ token is stored. The geometry terms of keypoints' the $n^{\text{th}}$ token are denoted as $\mathbf{K}_n=[\mathbf{k}_n^1,\mathbf{k}_n^2,...,\mathbf{k}_n^J]^\intercal$. We follow the hierarchical token scheme~\cite{zhang2021direct_mvp} to reduce the state search space, thus resulting in fewer learnable parameters.\\
\vspace{-0.75em}

\begin{figure*}[t]
    \centering
    \includegraphics[width=\textwidth]{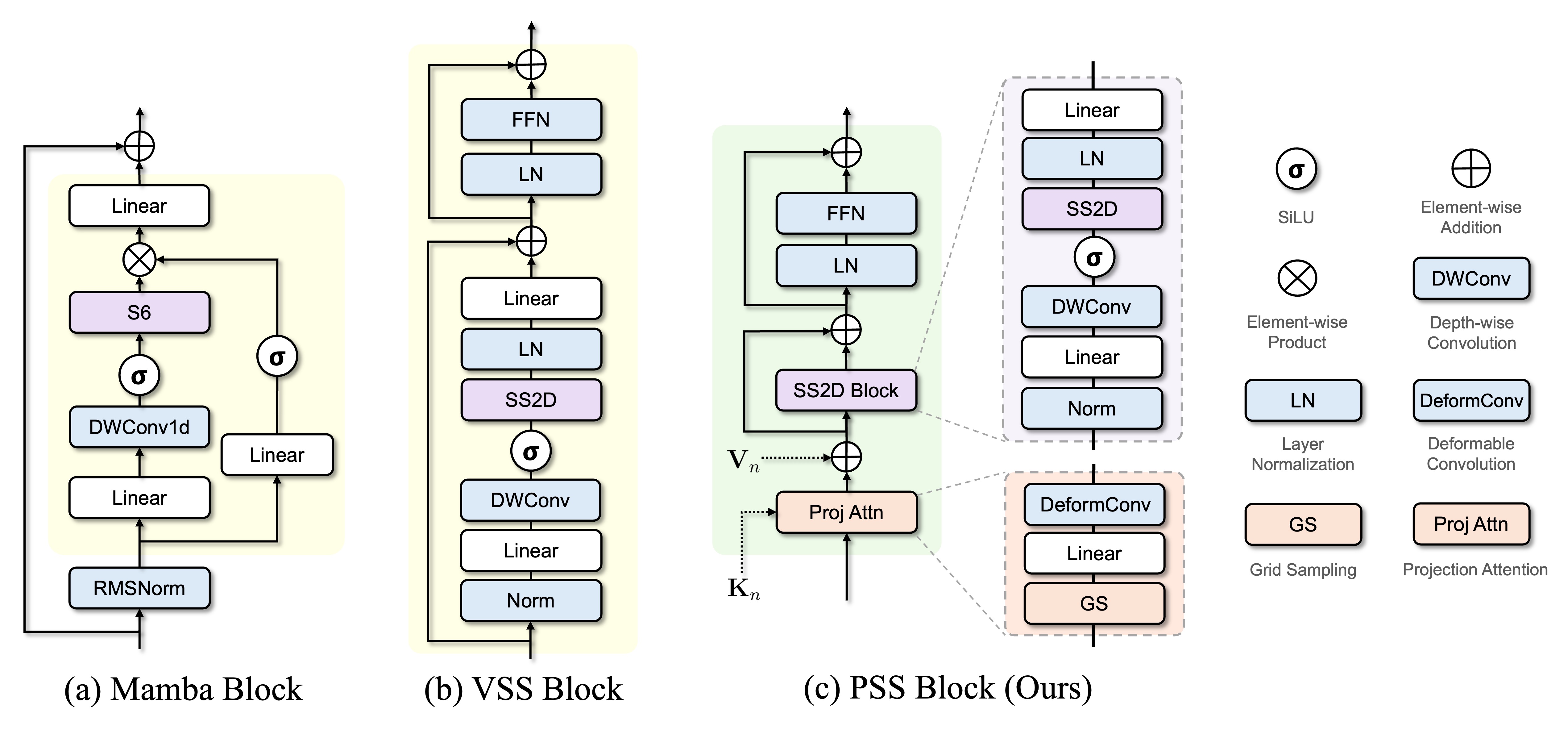}
    \vspace{-4mm}
    \caption{The architecture of Mamba block~\cite{gu2023mamba}, VSS block~\cite{liu2024vmamba} and the proposed Projective State Space block. The PSS block captures joint spatial relationships through projective attention and state-space modeling, progressively refining results.}
    \label{fig:pss_block}
\end{figure*}

\noindent \textbf{Projective State Space (PSS) Block.} A simple cross-attention on extracted tokens would extensively interact with the features across all spatial locations from multiple views. However, this leads to high computation costs and inefficiency due to multi-view features being involved. Comparatively, the projective attention~\cite{zhang2021direct_mvp} projects keypoints to anchor points onto different views and sample deformable points surrounding them to aggregate the local context. Thus, it encodes the corresponding 2D spatial positions across views and the local context extracted from sampling deformable points surrounding the anchors. This enables an efficient fusion of multi-view visual information and is adopted by several studies~\cite{zhang2021direct_mvp,liao2024multiple_mvgformer}. However, projective attention alone may not fully exploit inter-joint relationships from the joint spatial sequence $\mathbf{E}_n$. The PSS block addresses this by considering features from both projective attention and state-space modeling.\\
\vspace{-0.75em}

The PSS block operates on multi-scale tokens and focuses on learning the joint spatial sequence and inherent inter-joint relations. The architecture of the PSS block is shown in Figure~\ref{fig:model_architecture}. Each block is composed of an SS2D block, a projective attention, followed by a normalization, a feed-forward neural network, with a residual connection. The features in each view obtained from the projective attention are passed forward. \textit{MV-SSM} contains $M$ stacked PSS blocks. The PSS block not only leverages projection attention but also considers features from state space modeling. This  enables \textit{MV-SSM} to enhance performance with a few crucial tokens by learning effective features, shown in our ablation study~\ref{sec:ablation_study}. We further highlight the architectural difference between the Mamba~\cite{gu2023mamba}, VSS~\cite{liu2024vmamba}, and our proposed PSS block in Figure~\ref{fig:pss_block}.\\
\vspace{-0.75em}

As shown in Figure~\ref{fig:model_architecture}, we feed the visual feature term through the PSS block for feature update. For the subsequent layers, the PSS blocks update the tokens progressively and refine the prediction with offsets. The feature $\mathbf{F}_t$ is obtained from the sampling 2D positions $\mathbf{u}_t$. The MLP then learns the 2D residual offsets $\Delta\mathbf{u}_t$. The MLP also produces a confidence score $\mathbf{c}_t$. We obtain the new 3D position $\mathbf{k}'$ using differentiable algebraic triangulation~\cite{iskakov2019learnable}, formulated as follows:
\begin{equation}
    \mathbf{k}' = \text{AlgTriangulation}(\mathbf{u'}_t, \mathbf{c}_t, \mathbf{\Pi}_t),
\label{eq:triangulation}
\end{equation}
where $\mathbf{u'}_t$ is the 2D positions in $T$ views, $\mathbf{c}_t$ is corresponding confidence scores, and $\mathbf{\Pi}_t$ is the projection matrix for view $t$. As the model is end-to-end differentiable, the estimated confidence helps reduce 3D pose errors. Following ~\cite{zhang2021direct_mvp}, the 2D position is refined by adding the residual to the projection, with \(\Delta\mathbf{u}_t\) supervised by projected ground-truth 2D poses. This process is formulated as shown below:
\begin{equation}
 \mathbf{u'}_t = \mathbf{u}_t + \Delta\mathbf{u}_t .
\end{equation}
The updated token can better encode the joint spatial sequence of the corresponding humans in the scene. Passing the feature through the PSS block further improves 3D pose estimation, as the 2D pose prediction benefits from re-projecting the triangulated 3D pose, allowing it to extract more relevant features. For successive layers, the geometry term is directly updated with the estimated 3D keypoints.\\
\vspace{-0.75em}

\noindent \textbf{Grid Token-guided Bidirectional Scanning (GTBS).} To achieve robust 3D keypoint estimation, we design the Grid Token-guided Bi-directional Scanning for multi-view human 3D pose estimation. GTBS acts on the sampled projection tokens from each view. As the layers progressively refine keypoints, it scans more relevant features, boosting performance. A naive approach scans all image patches from multi-views (as shown in (a) of Figure~\ref{fig:motivation}), but it is computationally expensive due to redundant background tokens and ignores the spatial relationships of body keypoints. To address this, we perform token-level bidirectional scanning on sampled features instead of scanning all tokens. We adapt SS2D~\cite{liu2024vmamba} for grid token-guided bidirectional scanning, tailoring it to the joint spatial sequence.\\
\vspace{-0.75em}

Following~\cite{liao2024multiple_mvgformer}, we initialize the visual feature term from a standard normal distribution. For the geometry term, rather than relying on 3D predictions from a pre-trained model, we sample \(N\) human centers on the ground plane and initialize the keypoints as a conventional T-pose. Since the initial number of assumed tokens is kept large, we train a simple linear layer to predict a score for each token based on the visual feature term. This layer maps an \(L\)-dimensional feature to two scores, which are fed via a sigmoid function to output the positive and negative probabilities, respectively. Overall score is the average of scores across all keypoints, filtering tokens whose scores are smaller than $\epsilon$.\\
\vspace{-0.75em}

We train MV-SSM with a combined loss: a pose loss that supervises both the 2D/ 3D keypoints from positive tokens, and a cross-entropy loss for the MLP classifier distinguishing positive and negative tokens. This loss is applied at each layer of the progressive regression. For token-level ground truth assignment, we use the anchor-based matching strategy~\cite{zhou2019objects}. Following~\cite{liao2024multiple_mvgformer}, we assign ground-truth poses to the nearest token as ``positive," while unassigned tokens are considered ``negative." We denote initial geometry term as $\{\mathbf{K}_n^{(0)}\}_{n=1}^N$, serves as the anchor. The ground-truth keypoints for $Z$ humans are $\{\mathbf{H}_z\}_{z=1}^Z \subset\mathbb{R}^{J \times 3}$. The $W$ nearest tokens $\{\mathbf{K}_{z(w)}\}_{w=1}^W$ are assigned to each $\mathbf{K}_z$ based on their geometry term. The loss function is formulated as follows:
\vspace{-2mm}

{\small \begin{multline}
\mathcal{L}_{\text{pose}} = \sum _{w=1}^W \bigg(\mathcal{L}_1\big(\mathbf{K}_{z(w)}, \mathbf{H}_z\big) + 
\sum_{t=1}^T \mathcal{L}_1\big(\mathbf{\hat{U}}_{{z(w)},t}, \mathbf{U}_{z,t}\big)\bigg) ,
\end{multline}}

where we denote $\mathbf{\hat{U}}_{z(w),t}$ as the estimated 2D pose of the $z(w)^{\text{th}}$ token in the $t^{\text{th}}$ view, and $\mathbf{U}_{z,t}$ as the ground truth.
\section{Experiments and Results}

\noindent \textbf{Datasets.} We train MV-SSM on the popular CMU Panoptic~\cite{joo2015panoptic} dataset. To facilitate a direct comparison, same experimental settings as prior works~\cite{tu2020voxelpose, zhang2021direct_mvp, liao2024multiple_mvgformer} was used in our experiments. For generalizability evaluation, we employ two additional well-established benchmarks, Campus~\cite{belagiannis20143d} and Shelf~\cite{belagiannis20143d}. We do not fine-tune on these datasets, which enables us to assess realistic cross-dataset generalization performance.\\
\vspace{-0.75em}

\noindent \textbf{Evaluation Metrics.} Average Precision (AP) and Recall, along with Mean Per Joint Position Error (MPJPE) is used as the evaluation metric. For Campus and Shelf benchmarks, the Percentage of Correct Parts (PCP) is reported. The camera pose is assumed to be known and utilized, following common practice in the literature for this task. All baselines assume the camera poses are known.\\
\vspace{-0.75em} 

\noindent \textbf{Implementation Details.} For extracting image features from multi-view RGB input, ResNet-50 backbone pre-trained on the COCO dataset was used as in previous works~\cite{tu2020voxelpose, zhang2021direct_mvp, liao2024multiple_mvgformer}. The learning rate was $4e^{-4}$ and the model was trained for $40$ epochs with early stopping. Additionally, a confidence threshold of $\epsilon=0.1$ is applied layer-wise to filter redundant predictions~\cite{zhang2021direct_mvp}, followed by Non-Maximum Suppression (NMS) to refine the process further~\cite{liao2024multiple_mvgformer}. For additional implementation details, refer to the supplementary materials. \\
\vspace{-0.75em}

\subsection{Main Results}

\begin{figure}[t]
    \centering
    \includegraphics[width=0.48\textwidth]{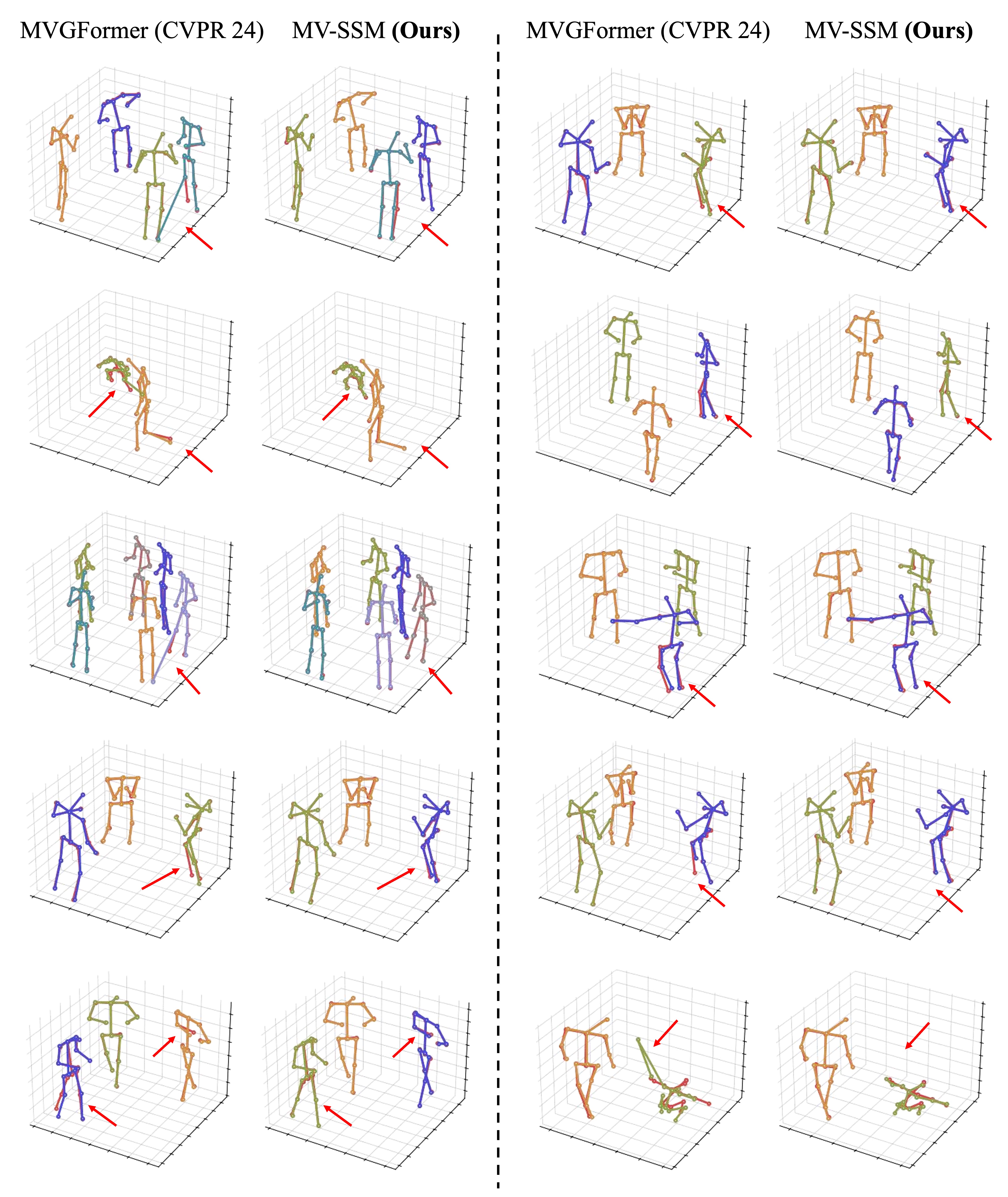}
    \caption{\textbf{Qualitative Comparisons.} We present a visual comparison against MVGFormer~\cite{liao2024multiple_mvgformer} on CMU Panoptic~\cite{joo2015panoptic} benchmark. The Ground truth human poses are shown in ``red'' and the predicted pose is overlapped on it to show an accurate comparison. MV-SSM achieves accurate poses, especially in difficult scenarios, demonstrating superior performance. As illustrated in the first row, MV-SSM is better able to predict, for example, the person's left foot. Note that the colors of persons are different since we do not perform ID-matching.}
    \label{fig:qualitative_comparison}
\end{figure}

\noindent \textbf{3D Keypoints Estimation Evaluation.} First, we evaluate MV-SSM's robustness in predicting 3D keypoints on the CMU Panoptic~\cite{joo2015panoptic} benchmark. The quantitative comparison with SOTAs is shown in Table~\ref{tab:in_domain}, demonstrating that MV-SSM significantly outperforms existing models. We use the same camera configurations and train-test splits as VoxelPose~\cite{tu2020voxelpose}, MvP~\cite{zhang2021direct_mvp}, and MVGFormer~\cite{liao2024multiple_mvgformer} for a fair comparison. MV-SSM surpasses other learning-based methods like VoxelPose by a large margin ({\color{red} $\mathbf{+9.5\%}$}) and further outperforms MvP and MVGFormer ({\color{red} $\mathbf{+1.2\%}$}). MvP tends to overfit training camera views and arrangements, thus lacking generalizability. We later demonstrate that MV-SSM surpasses both MvP and MVGFormer by a significant margin on generalizability. Qualitative visual comparisons with current SOTA~\cite{liao2024multiple_mvgformer} are shown in Figure~\ref{fig:qualitative_comparison}.

\vspace{2mm}
\noindent \textbf{Cross-dataset Generalizability Evaluation.} Next, we conduct a cross-scene evaluation by testing MV-SSM model trained on CMU-Panoptic directly on the Campus and Shelf benchmarks, without any additional fine-tuning. The comparative results are reported in Table~\ref{tab:cross_dataset}. As to Campus, MV-SSM demonstrates significant improvement compared to MVGFormer ({\color{red} $\mathbf{+15.3}$} on PCP for Actor 1 and {\color{red} $\mathbf{+9.2}$} PCP Average improvement), showcasing superior generalization capabilities. On the Shelf dataset, performance scores are almost saturated, and it is not surprising that the margin of improvement for our method is smaller. The lower PCP scores on Campus can be attributed to its more challenging nature--larger field-of-view and outdoor setting--compared to Shelf's controlled indoor environment. Note that geometry-based models generally achieve higher raw PCP scores than learning-based models. However, they struggle in complex occlusion scenarios due to reliance on 2D keypoints while neglecting feature-level visual information, leading to failures in matching humans across views. We reported the results without fine-tuning because fine-tuning does not provide a clear assessment of a model's transferability to new datasets.\\
\vspace{-0.75em}

\begin{table}[t]
\caption{\textbf{In-domain Results} on CMU Panoptic with conventional train-test split. `-' indicates information not reported by model.}
\centering
\resizebox{\columnwidth}{!}{%
\begin{tabular}{lc|cc}
\toprule
\multicolumn{1}{c}{\multirow{2}{*}{Method}} & \multicolumn{1}{c|}{\multirow{2}{*}{Venue}} & \multicolumn{2}{c}{CMU Panoptic} \\
\multicolumn{1}{c}{} & & AP$_{25}$ $\uparrow$ & MPJPE (mm) $\downarrow$ \\ 
\midrule
VoxelPose~\cite{tu2020voxelpose} & ECCV~20 & 84.0 & 17.7 \\
Lin \textit{et al.}~\cite{lin2021multi} & CVPR~21 & 92.1 & 16.8 \\
Graph3D~\cite{wu2021graph} & ICCV~21 & - & 15.8 \\
MvP~\cite{zhang2021direct_mvp} & NeurIPS~21 & 92.3 & 15.8 \\
Faster VoxelPose~\cite{ye2022faster} & ECCV~22 & 85.2 & 18.3 \\
MVGFormer~\cite{liao2024multiple_mvgformer} & CVPR~24 & 92.3 & 16.0 \\
\textbf{MV-SSM} & \textbf{Ours} & \cellcolor{green!40}\textbf{93.5} & \cellcolor{green!40}\textbf{15.7} \\
\bottomrule
\end{tabular}
}
\label{tab:in_domain}
\end{table}

\vspace{-2mm}
\begin{table*}[h]
\renewcommand{\arraystretch}{0.90}
\centering
\caption{\textbf{Cross-scene Generalization Evaluation.} All models were trained on CMU0 configuration of the CMU Panoptic benchmark~\cite{joo2015panoptic} and evaluated on Campus and Shelf~\cite{belagiannis20143d} w/o fine-tuning.}
\resizebox{1.60\columnwidth}{!}{%
\begin{tabular}{lc|ccccccccc@{}}
\toprule
\multirow{2}{*}{Method} & \multirow{2}{*}{Venue} & \multicolumn{4}{c|}{Campus (PCP$\uparrow$)} & \multicolumn{4}{c}{Shelf (PCP$\uparrow$)} \\
& & A1 & A2 & A3 & \multicolumn{1}{c|}{Average} & A1 & A2 & A3 & Average \\
\midrule
Dong \textit{et al.}~\cite{dong2019fast} & CVPR 19 & 97.6 & 93.3 & 98.0 & \multicolumn{1}{c|}{96.3} & 98.8 & 94.1 & 97.8 & 96.9 \\ 
\midrule
MvP~\cite{zhang2021direct_mvp} & NeurIPS 21 & 0.0 & 0.0 & 0.0 & \multicolumn{1}{c|}{0.0} & 5.7 & 16.0 & 4.6 & 8.7 \\
VoxelPose~\cite{tu2020voxelpose} & ECCV 20 & 16.1 & 12.4 & 5.1 & \multicolumn{1}{c|}{11.2} & 77.9 & 63.2 & 68.1 & 69.8 \\
MVGFormer~\cite{liao2024multiple_mvgformer} & CVPR 24 & 40.2 & 61.0 & 73.1 & \multicolumn{1}{c|}{58.1} & 89.9 & \cellcolor{green!40} \textbf{85.4} & 88.6 & 87.9\\
\textbf{MV-SSM} & \textbf{Ours} & \cellcolor{green!40}\textbf{55.5} & \cellcolor{green!40}\textbf{65.5} & \cellcolor{green!40}\textbf{79.9} & \multicolumn{1}{c|}{\cellcolor{green!40}\textbf{67.3}} & \cellcolor{green!40}\textbf{89.9} & 85.2 & \cellcolor{green!40}\textbf{88.9} & \cellcolor{green!40}\textbf{88.0}\\
\bottomrule
\end{tabular}
}
\setlength{\belowcaptionskip}{-4pt}
\label{tab:cross_dataset}
\end{table*}
\vspace{2mm}

\renewcommand{\arraystretch}{0.85}
\setlength{\tabcolsep}{5pt}
\begin{table*}[t]
\centering
\caption{\textbf{Cross-camera and Cross-arrangements Generalization Evaluation}. Cross-camera inference dataset CMU0 ($K$) indicates a total of $K$ cameras. In cross-arrangements, dataset CMU1-4 has varying numbers of cameras and camera IDs compared with CMU0. The CMU0 configuration with $5$ cameras is used for training all models, and results are reported w/o fine-tuning. Average is estimated over CMU0 (K) for K=$1$ to $7$.}

\resizebox{2\columnwidth}{!}{%
\begin{tabular}{@{}c|lc|cccccccccc}
 \toprule
& \multirow{2}{*}{Method} & \multirow{2}{*}{Venue} & \multicolumn{2}{c|}{CMU0 (3)} & \multicolumn{2}{c|}{CMU0 (4)} & \multicolumn{2}{c|}{CMU0 (6)} & \multicolumn{2}{c|}{CMU0 (7)} & \multicolumn{2}{c}{Average} \\
& & & AP$_{25}\uparrow$ & \multicolumn{1}{c|}{mAP$\uparrow$} & AP$_{25}$ & \multicolumn{1}{c|}{mAP$\uparrow$} & AP$_{25}\uparrow$ & \multicolumn{1}{c|}{mAP$\uparrow$} & AP$_{25}\uparrow$ & \multicolumn{1}{c|}{mAP$\uparrow$} & AP$_{25}\uparrow$ & mAP$\uparrow$ \\

\toprule
\parbox[t]{2mm}{\multirow{6}{*}{\rotatebox[origin=c]{90}{\small \ \ \ \ Cross-Camera}}} & Dong \textit{et al.}~\cite{dong2019fast} & CVPR~19 & 0.0 & \multicolumn{1}{c|}{37.7} & 0.0 & \multicolumn{1}{c|}{46.6} & 0.1 & \multicolumn{1}{c|}{41.8} & 0.2 & \multicolumn{1}{c|}{40.1} & 13.3 & 41.6 \\
& VoxelPose~\cite{tu2020voxelpose} & ECCV~20 & 13.7 & \multicolumn{1}{c|}{73.0} & 65.4 & \multicolumn{1}{c|}{93.1} & 84.5 & \multicolumn{1}{c|}{96.8} & 81.5 & \multicolumn{1}{c|}{96.3} & 69.1 & 89.8 \\
& MvP~\cite{zhang2021direct_mvp} & NeurIPS~21 & 12.3 & \multicolumn{1}{c|}{57.1} & 59.8 & \multicolumn{1}{c|}{84.1} & 0.0 & \multicolumn{1}{c|}{0.3} & 0.0 & \multicolumn{1}{c|}{0.2} & 21.6 & 35.4 \\
& MVGFormer~\cite{liao2024multiple_mvgformer} & CVPR~24 & 44.6 & \multicolumn{1}{c|}{83.4} & 83.6 & \multicolumn{1}{c|}{96.0} & \cellcolor{green!40}\textbf{94.7} & \multicolumn{1}{c|}{98.5} & 95.1 & \multicolumn{1}{c|}{98.6} & 83.3 & 94.1 \\
& \textbf{MV-SSM} & \textbf{Ours} & \cellcolor{green!40}\textbf{55.4} & \multicolumn{1}{c|}{\cellcolor{green!40}\textbf{90.3}} & \cellcolor{green!40}\textbf{84.9} & \multicolumn{1}{c|}{\cellcolor{green!40}\textbf{97.0}} & 94.6 & \multicolumn{1}{c|}{\cellcolor{green!40}\textbf{98.8}} & \cellcolor{green!40}\textbf{95.6} & \multicolumn{1}{c|}{\cellcolor{green!40}\textbf{98.8}} & \cellcolor{green!40}\textbf{84.8} & \cellcolor{green!40}\textbf{96.7} \\

\midrule
& Method & Venue & \multicolumn{2}{c|}{CMU1} & \multicolumn{2}{c|}{CMU2} & \multicolumn{2}{c|}{CMU3} & \multicolumn{2}{c|}{CMU4} & \multicolumn{2}{c}{Average} \\

\midrule
\parbox[t]{2mm}{\multirow{6}{*}{\rotatebox[origin=c]{90}{\small \ \ \ \ \ Cross-arrange}}} & Dong \textit{et al.}~\cite{dong2019fast} & CVPR~19 & 0.1 & \multicolumn{1}{c|}{39.9} & 0.0 & \multicolumn{1}{c|}{29.2} & 0.0 & \multicolumn{1}{c|}{29.2} & 0.0 & \multicolumn{1}{c|}{32.8} & 0.0 & 32.8 \\
& VoxelPose~\cite{tu2020voxelpose} & ECCV~20 & 28.8 & \multicolumn{1}{c|}{86.9} & 26.9 & \multicolumn{1}{c|}{69.6} & 21.0 & \multicolumn{1}{c|}{69.6} & 40.9 & \multicolumn{1}{c|}{87.9} & 29.4 & 78.5 \\
& MvP~\cite{zhang2021direct_mvp} & NeurIPS~21 & 0.0 & \multicolumn{1}{c|}{0.0} & 0.0 & \multicolumn{1}{c|}{0.0}  & 0.0 & \multicolumn{1}{c|}{0.0} & 0.0 & \multicolumn{1}{c|}{0.0} & 0.0 & 0.0 \\
& MVGFormer~\cite{liao2024multiple_mvgformer} & CVPR~24 & \cellcolor{green!40}\textbf{86.8} & \multicolumn{1}{c|}{96.1} & 67.9 & \multicolumn{1}{c|}{90.3} & 52.5 & \multicolumn{1}{c|}{78.8} & 91.5 & \multicolumn{1}{c|}{97.2} & 74.7 & 90.6 \\
& \textbf{MV-SSM} & \textbf{Ours} & 86.0 & \multicolumn{1}{c|}{\cellcolor{green!40}\textbf{97.0}} & \cellcolor{green!40}\textbf{74.6} & \multicolumn{1}{c|}{\cellcolor{green!40}\textbf{93.5}} & \cellcolor{green!40}\textbf{59.5} & \multicolumn{1}{c|}{\cellcolor{green!40}\textbf{84.8}} & \cellcolor{green!40}\textbf{94.2} & \multicolumn{1}{c|}{\cellcolor{green!40}\textbf{98.5}} & \cellcolor{green!40}\textbf{78.6} & \cellcolor{green!40}\textbf{93.5} \\ 
\bottomrule
\end{tabular}
}
\label{tab:cross-camera-arrangement}
\end{table*}

\noindent \textbf{Cross-camera Generalizability Evaluation.} Following the protocol in~\cite{bartol2022generalizable}, we evaluate MV-SSM's generalizability across varying numbers of cameras. When training the model we used the CMU0 camera setup with $5$ cameras (camera IDs: $3$, $6$, $12$, $13$, $23$). During testing, we systematically vary the number of cameras by adding or removing them from the CMU0 test set, following previous paradigms~\cite{liao2024multiple_mvgformer}. This configuration is denoted as ``CMU0($K$),'' where $K$ represents the number of cameras used during inference, as reported in Table~\ref{tab:cross-camera-arrangement}. We notice a consistent trend: as the number of cameras decreases to $4$ and then $3$, AP$_{25}$ and mAP scores drop. This performance drop is due to the loss of multi-view visual information from the removed cameras. Dong~\textit{et al.}~\cite{dong2019fast} fails to generate accurate 3D human keypoints, likely due to challenges in geometry-based cross-view matching. VoxelPose~\cite{tu2020voxelpose} exhibits a performance drop as the number of cameras increases, indicating overfitting to specific training camera configurations. Similarly, MvP~\cite{zhang2021direct_mvp} shows poor generalizability, as it overfits training cameras and fails on unseen camera numbers. Compared to MVGFormer~\cite{liao2024multiple_mvgformer}, our proposed model shows substantial improvement ({\color{red} $\mathbf{+10.8}$} on AP$_{25}$ and {\color{red}$\mathbf{+6.9}$} on mAP) in the challenging scenario when just using $3$ cameras. This suggests superior generalizability compared to MVGFormer. Furthermore, as the number of camera increases to $6$ and $7$, MV-SSM continues to surpass MVGFormer~\cite{liao2024multiple_mvgformer}, demonstrating its ability to effectively leverage additional visual information even for unseen camera views. In contrast, both MvP and VoxelPose demonstrate a performance drop as more cameras are added: MvP due to overfitting to specific camera arrangements during training, and VoxelPose due to not fully exploiting the visual information.\\
\vspace{-0.75em}

\noindent \textbf{Cross-arrangement Generalizability Evaluation.} We further increase the difficulty by conducting a cross-arrangement generalizability evaluation. In this setting, we change the camera arrangement, using different numbers of cameras and different camera IDs compared to the training set. The results are summarized in the lower half of Table~\ref{tab:cross-camera-arrangement}. The results demonstrate that MV-SSM consistently outperforms SOTA models, including the transformer-based MVGFormer. On the challenging CMU3, it achieves a high improvement of {\color{red} $\mathbf{+7.0}$} on AP$_{25}$ and {\color{red}$\mathbf{+6.0}$} on mAP compared to the current SOTA. Additionally, it significantly outperforms MvP, VoxelPose, and Dong \textit{et al.}~\cite{dong2019fast} thereby offering higher practical value in real-world scenarios where camera setups vary drastically. Its ability to generalize effectively across different camera arrangements w/o fine-tuning highlights its robustness and adaptability.

\subsection{Ablation Studies}
\label{sec:ablation_study}

\noindent \textbf{Effect of Proposed Modules.} The PSS block is a key contribution to our work, and we evaluate the effectiveness of each module within it. The results are presented in Table~\ref{tab:ablation}. PSS learns the joint spatial sequence and inter-joint relations using the state space model. In the first experiment, we replace the PSS block with a simple Mean block {\color{cvprblue}(Row 1)}, which is used to update the tokens by combining multi-view features. Since the mean operation will discard information, it is not surprising that it has a very low AP$_{25}$ and a high MPJPE. This result highlights the importance of encoding spatial and relational information. Next, we replace the PSS block with cross-attention {\color{cvprblue}(Row 2)}. The cross-attention alone cannot fully capture the complex dependencies modeled by our PSS block. A common objection is that the input features and Projective attention alone suffice for multi-view 3D pose estimation, with little benefit from the Mamba Blocks. To address this, we remove the Mamba (SS2D + LN + FFN) block from the model {\color{cvprblue}(Row 3)}, and the PSS block degenerates into a simple Projective attention. This results in a noticeable performance drop compared to the full model, confirming that the Mamba block is essential for capturing critical dependencies within PSS. A larger performance drop occurs when the proposed GTBS and Mamba blocks are removed {\color{cvprblue}(Row 4)}. Specifically, the AP$_{25}$ ($\uparrow$) drops to $93.5 \rightarrow 87.7$ while the MPJPE ($\downarrow$) error increases from $15.7 \rightarrow 18.6$. This demonstrates that both components are integral to achieving optimal performance. Table~\ref{tab:ablation} confirms the effectiveness of proposed modules.\\
\vspace{-0.75em}

\begin{table}
\renewcommand{\arraystretch}{0.90}
  \caption{\textbf{Ablation study} to verify proposed components. Results are reported on CMU Panoptic~\cite{joo2015panoptic}. Without is abbreviated as `w/o' which refers to the removed component / branch.}
   \vspace{-1mm}
  \label{tab:ablation}
  \centering
  \resizebox{0.49\textwidth}{!}{
    \begin{tabular}{c|l|ccc}
    \toprule
    & Ablation & AP$_{25}\uparrow$ & AP$_{100}$ $\uparrow$ & MPJPE $\downarrow$ \\
    \midrule
    \multicolumn{2}{l}{Module-wise Ablations} & & & \\
    \midrule
    1 & w Mean            & 36.2 & 54.0 & 71.8 \\
    2 & w Cross-attention & 90.4 & 99.1 & 16.8 \\
    3 & w/o Mamba (SS2D+LN+FFN) & 92.3 & 99.3 & 16.0 \\
    4 & w/o GTBS + Mamba & 87.7 & 98.8 & 18.6 \\
    \midrule
    \multicolumn{2}{l}{Branch-wise Ablations} & & & \\
    \midrule
    5 & w/o PSS\_$\mathbf{K}_n$\_Branch & 91.8 & 99.7 & 15.7 \\       
    6 & w/o PSS\_$\mathbf{V}_n$\_Branch & 92.5 & 99.6 & 15.9 \\
    \midrule
    \multicolumn{2}{l}{\bf MV-SSM (Full)} & \textbf{93.5} & \textbf{99.8}  &  \textbf{15.7} \\
    \bottomrule
  \end{tabular}
}
\vspace{-1em}
\end{table}

\noindent \textbf{Effect of Branch-wise Features.} 
As reported in Table~\ref{tab:ablation}, we validate each branch feature's contribution by systematically excluding them. First, we remove the $\mathbf{K}_n$ branch {\color{cvprblue}(Row 5)} and replace it with a simple MLP. This results in a significant performance drop, with AP$_{25}$ dropping from $93.5 \rightarrow 91.8$ and AP$_{100}$ decreasing from $99.8 \rightarrow 99.7$. This highlights the importance of explicitly modeling 3D keypoints to learn local context. Removing the $\mathbf{V}_n$ branch {\color{cvprblue}(Row 6)} causes a performance drop, though less important than without the 3D keypoints branch. This is because the 3D joints branch specifically learns strong local context. Here, AP$_{25}$ ($\uparrow$) value drops from $93.5 \rightarrow 92.5$ while the MPJPE ($\downarrow$) error increases from $15.7 \rightarrow 15.9$.

\section{Conclusion and Future Work}

The present work proposes \textit{MV-SSM}, a Multi-view State Space Model that enhances 3D human pose estimation by leveraging joint spatial sequences from multi-view inputs. \textit{MV-SSM} uses Projective State Space blocks to model joint spatial sequences and inter-joint relationships, achieving robust pose estimates. We introduce the progressive regression with Grid Token-guided Bidirectional Scan, enhancing both in- and out-of-domain performance and demonstrating strong generalizability to unseen cameras and geometries. Extensive experiments demonstrate that \textit{MV-SSM} outperforms previous state-of-the-art methods in generalizability, advancing multi-view 3D keypoint estimation for applications in AR/VR, animation, surveillance, etc.

Like previous models, \textit{MV-SSM} fails when a human is visible in only one camera, leading to missing keypoints from other views, especially when fewer cameras are involved. Future work could focus on extending the model to perform better in the single-view scenarios when the person is hidden from other views. Another potential avenue for improvement is enhancing robustness with generative models that can synthesize the missing keypoints or estimate the occluded keypoints, rather than randomly extracting them from a normal distribution. Future studies could also explore semi-supervised and self-supervised training strategies to leverage unlabeled multi-view images for augmenting the training datasets.

\vspace{0.5em}
\noindent \textbf{Acknowledgments.} Aviral Chharia was supported in part by the ATK-Nick G. Vlahakis Graduate Fellowship from Carnegie Mellon University, USA. \\
{
    \small
    \bibliographystyle{ieeenat_fullname}
    \bibliography{main}

\begin{thebibliography}{47}
\providecommand{\natexlab}[1]{#1}
\providecommand{\url}[1]{\texttt{#1}}
\expandafter\ifx\csname urlstyle\endcsname\relax
  \providecommand{\doi}[1]{doi: #1}\else
  \providecommand{\doi}{doi: \begingroup \urlstyle{rm}\Url}\fi

\bibitem[Akhter and Black(2015)]{akhter2015pose}
Ijaz Akhter and Michael~J Black.
\newblock Pose-conditioned joint angle limits for 3d human pose reconstruction.
\newblock In \emph{Proceedings of the IEEE conference on computer vision and pattern recognition}, pages 1446--1455, 2015.

\bibitem[Bartol et~al.(2022)Bartol, Bojani{\'c}, Petkovi{\'c}, and Pribani{\'c}]{bartol2022generalizable}
Kristijan Bartol, David Bojani{\'c}, Tomislav Petkovi{\'c}, and Tomislav Pribani{\'c}.
\newblock Generalizable human pose triangulation.
\newblock In \emph{Proceedings of the IEEE/CVF Conference on Computer Vision and Pattern Recognition}, pages 11028--11037, 2022.

\bibitem[Belagiannis et~al.(2014)Belagiannis, Amin, Andriluka, Schiele, Navab, and Ilic]{belagiannis20143d}
Vasileios Belagiannis, Sikandar Amin, Mykhaylo Andriluka, Bernt Schiele, Nassir Navab, and Slobodan Ilic.
\newblock 3d pictorial structures for multiple human pose estimation.
\newblock In \emph{Proceedings of the IEEE conference on computer vision and pattern recognition}, pages 1669--1676, 2014.

\bibitem[Bridgeman et~al.(2019)Bridgeman, Volino, Guillemaut, and Hilton]{bridgeman2019multi}
Lewis Bridgeman, Marco Volino, Jean-Yves Guillemaut, and Adrian Hilton.
\newblock Multi-person 3d pose estimation and tracking in sports.
\newblock In \emph{Proceedings of the IEEE/CVF conference on computer vision and pattern recognition workshops}, pages 0--0, 2019.

\bibitem[Cao et~al.(2019)Cao, Hidalgo, Simon, Wei, and Sheikh]{cao2019openposerealtimemultiperson2d}
Zhe Cao, Gines Hidalgo, Tomas Simon, Shih-En Wei, and Yaser Sheikh.
\newblock Openpose: Realtime multi-person 2d pose estimation using part affinity fields, 2019.

\bibitem[Chen et~al.(2020{\natexlab{a}})Chen, Guo, Li, Lee, and Chirikjian]{chen2020multi}
He Chen, Pengfei Guo, Pengfei Li, Gim~Hee Lee, and Gregory Chirikjian.
\newblock Multi-person 3d pose estimation in crowded scenes based on multi-view geometry.
\newblock In \emph{Computer Vision--ECCV 2020: 16th European Conference, Glasgow, UK, August 23--28, 2020, Proceedings, Part III 16}, pages 541--557. Springer, 2020{\natexlab{a}}.

\bibitem[Chen et~al.(2020{\natexlab{b}})Chen, Ai, Chen, Zhuang, and Liu]{chen2020cross}
Long Chen, Haizhou Ai, Rui Chen, Zijie Zhuang, and Shuang Liu.
\newblock Cross-view tracking for multi-human 3d pose estimation at over 100 fps.
\newblock In \emph{Proceedings of the IEEE/CVF conference on computer vision and pattern recognition}, pages 3279--3288, 2020{\natexlab{b}}.

\bibitem[Chu et~al.(2021)Chu, Lee, Lee, Hsu, Li, and Chen]{chu2021part}
Hau Chu, Jia-Hong Lee, Yao-Chih Lee, Ching-Hsien Hsu, Jia-Da Li, and Chu-Song Chen.
\newblock Part-aware measurement for robust multi-view multi-human 3d pose estimation and tracking.
\newblock In \emph{Proceedings of the IEEE/CVF Conference on Computer Vision and Pattern Recognition}, pages 1472--1481, 2021.

\bibitem[Dabral et~al.(2018)Dabral, Mundhada, Kusupati, Afaque, Sharma, and Jain]{dabral2018learning}
Rishabh Dabral, Anurag Mundhada, Uday Kusupati, Safeer Afaque, Abhishek Sharma, and Arjun Jain.
\newblock Learning 3d human pose from structure and motion.
\newblock In \emph{Proceedings of the European conference on computer vision (ECCV)}, pages 668--683, 2018.

\bibitem[Defferrard et~al.(2016)Defferrard, Bresson, and Vandergheynst]{defferrard2016convolutional}
Micha{\"e}l Defferrard, Xavier Bresson, and Pierre Vandergheynst.
\newblock Convolutional neural networks on graphs with fast localized spectral filtering.
\newblock \emph{Advances in neural information processing systems}, 29, 2016.

\bibitem[Deng et~al.(2009)Deng, Dong, Socher, Li, Li, and Fei-Fei]{deng2009imagenet}
Jia Deng, Wei Dong, Richard Socher, Li-Jia Li, Kai Li, and Li Fei-Fei.
\newblock Imagenet: A large-scale hierarchical image database.
\newblock In \emph{2009 IEEE conference on computer vision and pattern recognition}, pages 248--255. Ieee, 2009.

\bibitem[Dong et~al.(2019)Dong, Jiang, Huang, Bao, and Zhou]{dong2019fast}
Junting Dong, Wen Jiang, Qixing Huang, Hujun Bao, and Xiaowei Zhou.
\newblock Fast and robust multi-person 3d pose estimation from multiple views.
\newblock In \emph{Proceedings of the IEEE/CVF conference on computer vision and pattern recognition}, pages 7792--7801, 2019.

\bibitem[Dong et~al.(2021)Dong, Fang, Jiang, Yang, Huang, Bao, and Zhou]{dong2021fast}
Junting Dong, Qi Fang, Wen Jiang, Yurou Yang, Qixing Huang, Hujun Bao, and Xiaowei Zhou.
\newblock Fast and robust multi-person 3d pose estimation and tracking from multiple views.
\newblock \emph{IEEE transactions on pattern analysis and machine intelligence}, 44\penalty0 (10):\penalty0 6981--6992, 2021.

\bibitem[Fan et~al.(2014)Fan, Zheng, Zhou, and Wang]{fan2014pose}
Xiaochuan Fan, Kang Zheng, Youjie Zhou, and Song Wang.
\newblock Pose locality constrained representation for 3d human pose reconstruction.
\newblock In \emph{Computer Vision--ECCV 2014: 13th European Conference, Zurich, Switzerland, September 6-12, 2014, Proceedings, Part I 13}, pages 174--188. Springer, 2014.

\bibitem[Gu and Dao(2023)]{gu2023mamba}
Albert Gu and Tri Dao.
\newblock Mamba: Linear-time sequence modeling with selective state spaces.
\newblock 2023.

\bibitem[Gu et~al.(2021{\natexlab{a}})Gu, Goel, and R{\'e}]{gu2021s4}
Albert Gu, Karan Goel, and Christopher R{\'e}.
\newblock Efficiently modeling long sequences with structured state spaces.
\newblock 2021{\natexlab{a}}.

\bibitem[Gu et~al.(2021{\natexlab{b}})Gu, Johnson, Goel, Saab, Dao, Rudra, and R{\'e}]{gu2021combining}
Albert Gu, Isys Johnson, Karan Goel, Khaled Saab, Tri Dao, Atri Rudra, and Christopher R{\'e}.
\newblock Combining recurrent, convolutional, and continuous-time models with linear state space layers.
\newblock \emph{Advances in neural information processing systems}, 34:\penalty0 572--585, 2021{\natexlab{b}}.

\bibitem[Hatamizadeh and Kautz(2024)]{hatamizadeh2024mambavisionhybridmambatransformervision}
Ali Hatamizadeh and Jan Kautz.
\newblock Mambavision: A hybrid mamba-transformer vision backbone, 2024.

\bibitem[He et~al.(2016)He, Zhang, Ren, and Sun]{he2016deep}
Kaiming He, Xiangyu Zhang, Shaoqing Ren, and Jian Sun.
\newblock Deep residual learning for image recognition.
\newblock In \emph{Proceedings of the IEEE conference on computer vision and pattern recognition}, pages 770--778, 2016.

\bibitem[Iskakov et~al.(2019)Iskakov, Burkov, Lempitsky, and Malkov]{iskakov2019learnable}
Karim Iskakov, Egor Burkov, Victor Lempitsky, and Yury Malkov.
\newblock Learnable triangulation of human pose.
\newblock In \emph{Proceedings of the IEEE/CVF international conference on computer vision}, pages 7718--7727, 2019.

\bibitem[Joo et~al.(2015)Joo, Liu, Tan, Gui, Nabbe, Matthews, Kanade, Nobuhara, and Sheikh]{joo2015panoptic}
Hanbyul Joo, Hao Liu, Lei Tan, Lin Gui, Bart Nabbe, Iain Matthews, Takeo Kanade, Shohei Nobuhara, and Yaser Sheikh.
\newblock Panoptic studio: A massively multiview system for social motion capture.
\newblock In \emph{Proceedings of the IEEE international conference on computer vision}, pages 3334--3342, 2015.

\bibitem[Kalman(1960)]{kalman1960new}
Rudolph~Emil Kalman.
\newblock A new approach to linear filtering and prediction problems.
\newblock 1960.

\bibitem[Li et~al.(2025{\natexlab{a}})Li, Li, Wang, He, Wang, Wang, and Qiao]{li2025videomamba}
Kunchang Li, Xinhao Li, Yi Wang, Yinan He, Yali Wang, Limin Wang, and Yu Qiao.
\newblock Videomamba: State space model for efficient video understanding.
\newblock In \emph{European Conference on Computer Vision}, pages 237--255. Springer, 2025{\natexlab{a}}.

\bibitem[Li and Chan(2015)]{li20153d}
Sijin Li and Antoni~B Chan.
\newblock 3d human pose estimation from monocular images with deep convolutional neural network.
\newblock In \emph{Computer Vision--ACCV 2014: 12th Asian Conference on Computer Vision, Singapore, Singapore, November 1-5, 2014, Revised Selected Papers, Part II 12}, pages 332--347. Springer, 2015.

\bibitem[Li et~al.(2025{\natexlab{b}})Li, Singh, and Grover]{li2025mamba}
Shufan Li, Harkanwar Singh, and Aditya Grover.
\newblock Mamba-nd: Selective state space modeling for multi-dimensional data.
\newblock In \emph{European Conference on Computer Vision}, pages 75--92. Springer, 2025{\natexlab{b}}.

\bibitem[Li et~al.(2022)Li, Liu, Tang, Wang, and Van~Gool]{Li_2022_CVPR}
Wenhao Li, Hong Liu, Hao Tang, Pichao Wang, and Luc Van~Gool.
\newblock Mhformer: Multi-hypothesis transformer for 3d human pose estimation.
\newblock In \emph{Proceedings of the IEEE/CVF Conference on Computer Vision and Pattern Recognition (CVPR)}, pages 13147--13156, 2022.

\bibitem[Liao et~al.(2024)Liao, Zhu, Wang, Hu, and Waslander]{liao2024multiple_mvgformer}
Ziwei Liao, Jialiang Zhu, Chunyu Wang, Han Hu, and Steven~L Waslander.
\newblock Multiple view geometry transformers for 3d human pose estimation.
\newblock In \emph{Proceedings of the IEEE/CVF Conference on Computer Vision and Pattern Recognition}, pages 708--717, 2024.

\bibitem[Lin and Lee(2021)]{lin2021multi}
Jiahao Lin and Gim~Hee Lee.
\newblock Multi-view multi-person 3d pose estimation with plane sweep stereo.
\newblock In \emph{Proceedings of the IEEE/CVF Conference on Computer Vision and Pattern Recognition}, pages 11886--11895, 2021.

\bibitem[Liu et~al.(2024)Liu, Tian, Zhao, Yu, Xie, Wang, Ye, and Liu]{liu2024vmamba}
Yue Liu, Yunjie Tian, Yuzhong Zhao, Hongtian Yu, Lingxi Xie, Yaowei Wang, Qixiang Ye, and Yunfan Liu.
\newblock Vmamba: Visual state space model.
\newblock 2024.

\bibitem[Ma et~al.(2021)Ma, Su, Wang, Ci, and Wang]{ma2021context}
Xiaoxuan Ma, Jiajun Su, Chunyu Wang, Hai Ci, and Yizhou Wang.
\newblock Context modeling in 3d human pose estimation: A unified perspective.
\newblock In \emph{Proceedings of the IEEE/CVF conference on computer vision and pattern recognition}, pages 6238--6247, 2021.

\bibitem[Martinez et~al.(2017)Martinez, Hossain, Romero, and Little]{martinez2017simple}
Julieta Martinez, Rayat Hossain, Javier Romero, and James~J Little.
\newblock A simple yet effective baseline for 3d human pose estimation.
\newblock In \emph{Proceedings of the IEEE international conference on computer vision}, pages 2640--2649, 2017.

\bibitem[Pavlakos et~al.(2017)Pavlakos, Zhou, Derpanis, and Daniilidis]{pavlakos2017coarse}
Georgios Pavlakos, Xiaowei Zhou, Konstantinos~G Derpanis, and Kostas Daniilidis.
\newblock Coarse-to-fine volumetric prediction for single-image 3d human pose.
\newblock In \emph{Proceedings of the IEEE conference on computer vision and pattern recognition}, pages 7025--7034, 2017.

\bibitem[Perez-Yus and Agudo(2022)]{perez2022matching}
Alejandro Perez-Yus and Antonio Agudo.
\newblock Matching and recovering 3d people from multiple views.
\newblock In \emph{Proceedings of the IEEE/CVF Winter Conference on Applications of Computer Vision}, pages 3622--3631, 2022.

\bibitem[Ramakrishna et~al.(2012)Ramakrishna, Kanade, and Sheikh]{ramakrishna2012reconstructing}
Varun Ramakrishna, Takeo Kanade, and Yaser Sheikh.
\newblock Reconstructing 3d human pose from 2d image landmarks.
\newblock In \emph{Computer Vision--ECCV 2012: 12th European Conference on Computer Vision, Florence, Italy, October 7-13, 2012, Proceedings, Part IV 12}, pages 573--586. Springer, 2012.

\bibitem[Reddy et~al.(2021)Reddy, Guigues, Pishchulin, Eledath, and Narasimhan]{reddy2021tessetrack}
N~Dinesh Reddy, Laurent Guigues, Leonid Pishchulin, Jayan Eledath, and Srinivasa~G Narasimhan.
\newblock Tessetrack: End-to-end learnable multi-person articulated 3d pose tracking.
\newblock In \emph{Proceedings of the IEEE/CVF conference on computer vision and pattern recognition}, pages 15190--15200, 2021.

\bibitem[Tu et~al.(2020)Tu, Wang, and Zeng]{tu2020voxelpose}
Hanyue Tu, Chunyu Wang, and Wenjun Zeng.
\newblock Voxelpose: Towards multi-camera 3d human pose estimation in wild environment.
\newblock In \emph{Computer Vision--ECCV 2020: 16th European Conference, Glasgow, UK, August 23--28, 2020, Proceedings, Part I 16}, pages 197--212. Springer, 2020.

\bibitem[Vaswani(2017)]{vaswani2017attention}
A Vaswani.
\newblock Attention is all you need.
\newblock \emph{Advances in Neural Information Processing Systems}, 2017.

\bibitem[Wang et~al.(2024)Wang, Chen, Liu, Chen, Lin, Han, and Ding]{wang2024yolov10}
Ao Wang, Hui Chen, Lihao Liu, Kai Chen, Zijia Lin, Jungong Han, and Guiguang Ding.
\newblock Yolov10: Real-time end-to-end object detection.
\newblock \emph{arXiv preprint arXiv:2405.14458}, 2024.

\bibitem[Wang et~al.(2014)Wang, Wang, Lin, Yuille, and Gao]{wang2014robust}
Chunyu Wang, Yizhou Wang, Zhouchen Lin, Alan~L Yuille, and Wen Gao.
\newblock Robust estimation of 3d human poses from a single image.
\newblock In \emph{Proceedings of the IEEE conference on computer vision and pattern recognition}, pages 2361--2368, 2014.

\bibitem[Wu et~al.(2021)Wu, Jin, Liu, Bai, Qian, Liu, and Ouyang]{wu2021graph}
Size Wu, Sheng Jin, Wentao Liu, Lei Bai, Chen Qian, Dong Liu, and Wanli Ouyang.
\newblock Graph-based 3d multi-person pose estimation using multi-view images.
\newblock In \emph{Proceedings of the IEEE/CVF international conference on computer vision}, pages 11148--11157, 2021.

\bibitem[Xiao et~al.(2018)Xiao, Wu, and Wei]{xiao2018simple}
Bin Xiao, Haiping Wu, and Yichen Wei.
\newblock Simple baselines for human pose estimation and tracking.
\newblock In \emph{Proceedings of the European conference on computer vision (ECCV)}, pages 466--481, 2018.

\bibitem[Yasin et~al.(2016)Yasin, Iqbal, Kruger, Weber, and Gall]{yasin2016dual}
Hashim Yasin, Umar Iqbal, Bjorn Kruger, Andreas Weber, and Juergen Gall.
\newblock A dual-source approach for 3d pose estimation from a single image.
\newblock In \emph{Proceedings of the IEEE Conference on Computer Vision and Pattern Recognition}, pages 4948--4956, 2016.

\bibitem[Ye et~al.(2022)Ye, Zhu, Wang, Wu, and Wang]{ye2022faster}
Hang Ye, Wentao Zhu, Chunyu Wang, Rujie Wu, and Yizhou Wang.
\newblock Faster voxelpose: Real-time 3d human pose estimation by orthographic projection.
\newblock In \emph{European Conference on Computer Vision}, pages 142--159. Springer, 2022.

\bibitem[Zhang et~al.(2021)Zhang, Cai, Yan, Feng, et~al.]{zhang2021direct_mvp}
Jianfeng Zhang, Yujun Cai, Shuicheng Yan, Jiashi Feng, et~al.
\newblock Direct multi-view multi-person 3d pose estimation.
\newblock \emph{Advances in Neural Information Processing Systems}, 34:\penalty0 13153--13164, 2021.

\bibitem[Zhang et~al.(2020)Zhang, An, Yu, Li, Li, and Liu]{zhang20204d}
Yuxiang Zhang, Liang An, Tao Yu, Xiu Li, Kun Li, and Yebin Liu.
\newblock 4d association graph for realtime multi-person motion capture using multiple video cameras.
\newblock In \emph{Proceedings of the IEEE/CVF conference on computer vision and pattern recognition}, pages 1324--1333, 2020.

\bibitem[Zhou et~al.(2019)Zhou, Wang, and Kr{\"a}henb{\"u}hl]{zhou2019objects}
Xingyi Zhou, Dequan Wang, and Philipp Kr{\"a}henb{\"u}hl.
\newblock Objects as points.
\newblock \emph{arXiv preprint arXiv:1904.07850}, 2019.

\bibitem[Zhu et~al.(2024)Zhu, Liao, Zhang, Wang, Liu, and Wang]{vim}
Lianghui Zhu, Bencheng Liao, Qian Zhang, Xinlong Wang, Wenyu Liu, and Xinggang Wang.
\newblock Vision mamba: Efficient visual representation learning with bidirectional state space model.
\newblock \emph{arXiv preprint arXiv:2401.09417}, 2024.

\end{thebibliography}
}
\clearpage
\setcounter{section}{0}
\setcounter{table}{0}
\maketitlesupplementary

\noindent The supplementary includes additional model architecture and training details in Section~\ref{supp:model_architecture_details}) and Section~\ref{supp:model_training}) respectively. We include in-depth details of the datasets (Section~\ref{supp:datasets}), training schemes (Section~\ref{supp:training_schemes}), and various evaluation metric definitions (Section~\ref{supp:evaluation_metrics}) used in the study. Additional details of the ablation studies are included in Section~\ref{supp:additional_details}.

\section{Model Architecture Details}
\label{supp:model_architecture_details}

The training images, sized $960 \times 512$, from the CMU-Panoptic dataset were fed into a ResNet-50 backbone pre-trained on the COCO dataset for 2D pose estimation task~\cite{xiao2018simple}. We used the same backbone weights as previous works~\cite{tu2020voxelpose, zhang2021direct_mvp, liao2024multiple_mvgformer} for a fair comparison. The model utilized $4$ PSS/ decoder layers. The state space (SS2D + LN + FFN) blocks in the proposed PSS blocks had a $256$-dimensional token size with a depth of $1$ and $2$ respectively. The token dimension is $256$. The gate control was deprecated, and downsampling was set to `none,' ensuring that the output shape of the state space blocks is the same as the input. It is important to note that the decoder layers did not share the parameters. Tokens with scores below $\epsilon=0.1$ were filtered out at each decoder layer~\cite{zhang2021direct_mvp}, followed by NMS to remove redundant tokens~\cite{liao2024multiple_mvgformer}. During initialization, the token number was approximately set to $1024$, based on the motion capture space of the dataset.

\section{Training and other details}
\label{supp:model_training}

\subsection{Datasets}
\label{supp:datasets}

We outline the details of the datasets employed in this section. These include the CMU Panoptic~\cite{joo2015panoptic}, Shelf~\cite{belagiannis20143d}, and Campus~\cite{belagiannis20143d} datasets. Note that only CMU Panoptic was used for training.

\begin{itemize}
    \item \textbf{CMU Panoptic}~\cite{joo2015panoptic} is a 3D multi-view dataset that contains multiple persons. The videos are collected in a spherical dome, i.e., an indoor scenario, where $480$ RGB cameras and $10$ RGBD cameras are installed. The Panoptic dataset contains over $30$ videos and $65$ sequences, including a variety of subjects wearing casual clothes, and performing a wide range of activities like dancing, playing musical instruments, eating, and so forth. The dataset is widely used in single and multi-view pose estimation tasks. Besides the 3D skeleton and point cloud labels, CMU Panoptic also provides facial landmarks, transcripts, and speaker ID, making it also suitable for whole-body or multi-modal tasks. 
    
    \item \textbf{Campus}~\cite{belagiannis20143d} dataset includes videos taken on a campus, for up to $3$ subjects performing various actions, from $3$ different views. The key points of subjects are annotated manually in the dataset.
    
    \item \textbf{Shelf}~\cite{belagiannis20143d} dataset, as it is named, contains videos from $5$ views of up to $4$ subjects disassembling a shelf and interacting with each other. The dataset comes with manually labeled keypoints in all views.
\end{itemize}

\subsection{Training Schemes}
\label{supp:training_schemes}

MV-SSM was trained on eight NVIDIA TITAN RTX GPUs with a batch size of $1$ for $40$ epochs using a learning rate of \texttt{4e-4}. The training process required around $1.5$ days. Early stopping was employed to prevent overfitting, and the backbone was kept frozen throughout training. For training on the CMU Panoptic dataset, the space size was set to $[8000, 8000, 2000]$, with the space center positioned at $[0.0, -500, 800]$. The initial cube size was set to $[80, 80, 20]$. For both the Campus and Shelf datasets, the space size remained constant, while the space centers was set at $[2000, 5000, 1000]$ for the Campus, and $[0, 500, 800]$ for the Shelf dataset.

\begin{table}[]
\centering
\caption{The details of the camera IDs, arrangements, and the camera numbers used in various experiments.}
\resizebox{\columnwidth}{!}{%
\begin{tabular}{llc}
\toprule
Cam Arrangements & Cam IDs & Numbers \\ 
\midrule
CMU1 & 1, 2, 3, 4, 6, 7, 10 & 7 \\
CMU2 & 12, 16, 18, 19, 22, 23, 30 & 7 \\
CMU3 & 10, 12, 16, 18 & 4 \\
CMU4 & 6, 7, 10, 12, 16, 18, 19, 22, 23, 30 & 10 \\ 
\midrule
CMU0 & 3, 6, 12, 13, 23 & 5 \\
CMU0 w/ 2 Extra & 3, 6, 12, 13, 23, 10, 16 & 7 \\
CMU0($K$) & First $K$ cameras in CMU0 w/ 2 & $K$ \\ 
\bottomrule
\end{tabular}
}
\label{table:camera_arrangements}
\end{table}

\subsection{Evaluation Metrics}
\label{supp:evaluation_metrics}

In this section, we provide a detailed definition of the evaluation metrics used in this study.

\begin{itemize}
    \item \textbf{MPJPE} or the Mean Per Joint Position Error is the average error in keypoint positions between the model's predictions and the corresponding ground truth. It is calculated as the Euclidean distance between the predicted keypoints and the ground truth keypoints. MPJPE is expressed in millimeters (mm) and is commonly used as the primary evaluation metric in 3D human pose estimation tasks.

    \item \textbf{AP and mAP} Average Precision (AP) is a widely used metric in various tasks, including classification and object detection. Average Precision (AP) is calculated as shown in Equation~\ref{eq:AP}, 
    \begin{equation}
        AP= \sum^{k =n-1}_{k=0}{[r(k)-r(k+1)]*p(k)}
        \label{eq:AP}
    \end{equation}
    where~$r(k)$ and~$p(k)$ denote the recall and precision at the~$k^{th}$ sample, respectively. AP is used to summarize predictions in a binary manner, i.e., whether the predicted 3D keypoints have an MPJPE below a certain threshold or not, to measure the overall prediction's accuracy within a given margin. The threshold is indicated after AP, e.g. $AP_{25}$ refers to the average precision value with an MPJPE threshold of 25mm. Mean Average Precision~(mAP) is the mean value of AP across multiple thresholds. We follow the previous works~\cite{zhang2021direct_mvp,liao2024multiple_mvgformer} and calculate mAP across MPJPE thresholds of $[25,50,75,100,125,150]$ mm.

    \item \textbf{PCP} stands for the Percentage of Correct Parts. This metric measures the Euclidean distance between the predicted endpoints of each limb and the ground truth.  If the average error between the two endpoints is less than half of the limb's length, that part of the body is considered correctly predicted. PCP is expressed as the percentage of correctly predicted parts out of the total number of parts.
\end{itemize}

\section{Ablation Details}
\label{supp:additional_details}

We provide an in-depth description of the modifications made to MV-SSM to perform the ablation experiments. Since the results of the ablations have already been discussed in detail in the main paper in Rows 1-6 of Table~4, we focus on an experiment-wise detailed description. The first four sets of experiments involve component-wise ablations, which help to test the how effective each of the proposed PSS block components were, while the remaining two discuss the branch-wise ablations by systematically excluding them from the model. We include both the component-wise and the branch-wise ablations to help establish the contribution of each proposed design improvement at both component and branch levels.

\begin{itemize}
    \item \textbf{w Mean.} In the first ablation, we remove the PSS Block from MV-SSM and replace it with a simple Mean Block {\color{cvprblue}(Row 1)}. Note that we still keep the Projective attention in the first block. Therefore, instead of the PSS Block, the Mean is used to update the tokens by averaging the multi-view features. Note that when the PSS block is removed, the multi-scale features from the backbone are directly input in the projective attention, and the resulting token is input into the mean block (instead of the PSS block). Note that the model still learns the projective attention features from the first layer, however, when it fuses the multi-view information, it discards this information leading to a drop in performance. However, the mean operation acts as a very naive baseline, and we simply use it to confirm the importance of encoding spatial and relational information, which were explicitly modeled by the PSS block but were discarded by the introduced mean block.
    
    \item \textbf{w Cross-attention.} Since the mean is a very naive baseline, we replace the PSS Block with cross-attention {\color{cvprblue}(Row 2)}. Note that for a fair comparison, the same architectural setting was followed as the previous ablation, where the model still learns the projective attention tokens. In this way, when using cross-attention, the model retains the feature information over subsequent layers.
    
    \item \textbf{w/o Mamba (SS2D + LN + FFN).} In the third ablation, to study the contribution of state space modeling, we remove the SS2D + LN + FFN blocks from MV-SSM {\color{cvprblue}(See Row 3)}. In doing so, the PSS block degenerates into a simple Projective attention. Note that this significantly differs from the previous ablations since the multi-view feature fusion is still performed by the degenerate PSS block (`that consists of the Proj Attn'), while in the previous ablations, it was being performed by the `mean' and `cross-attention'.

    \item \textbf{w/o GTBS + Mamba.} In the fourth ablation, we remove the GT-Bidirectional Scan and the Mamba blocks. For this, we modified the appearance token to encode only the instance-level information and removed the Mamba blocks. In this way, only the instance-level tokens are scanned. Note that the multi-view feature fusion also degenerates to only fuse the instance level features {\color{cvprblue}(Row 4)}.

    \item \textbf{w/o PSS\_$\mathbf{K}_n$\_Branch.} In the branch-wise ablation, we remove the geometric token update branch (or simply the 3D keypoints branch) {\color{cvprblue}(Row 5)} and replace it with a simple MLP.

    \item \textbf{w/o PSS\_$\mathbf{V}_n$\_Branch.} For removing the visual token update branch (or simply visual feature branch) {\color{cvprblue}(Row 6)}.
\end{itemize}
\end{document}